\documentclass[conference]{IEEEtran}
\IEEEoverridecommandlockouts
\usepackage{cite}
\usepackage{amsmath,amssymb,amsfonts}
\usepackage{algorithmic}
\usepackage{graphicx}
\usepackage{textcomp}
\usepackage{xcolor}

\newcommand{\ignore}[1]{}
\newcommand{\Name}{EmoXpt}
\def\BibTeX{{\rm B\kern-.05em{\sc i\kern-.025em b}\kern-.08em
    T\kern-.1667em\lower.7ex\hbox{E}\kern-.125emX}}
\begin{document}

\title{\Name: Analyzing Emotional Variances in Human Comments and LLM-Generated Responses}


\author{\IEEEauthorblockN{Shireesh Reddy Pyreddy}
\IEEEauthorblockA{\textit{Department of Computer Science} \\
\textit{SUNY Polytechnic Institute}\\
Utica, NY, USA \\
pyredds@sunypoly.edu}
\and
\IEEEauthorblockN{Tarannum Shaila Zaman}
\IEEEauthorblockA{\textit{Department of Information Systems} \\
\textit{University of Maryland Baltimore County}\\
Baltimore, MD, USA \\
zamant@umbc.edu}
}

\maketitle

\begin{abstract}

The widespread adoption of generative AI has generated diverse opinions, with individuals expressing both support and criticism of its applications. This study investigates the emotional dynamics surrounding generative AI by analyzing human tweets referencing terms such as ChatGPT, OpenAI, Copilot, and LLMs. To further understand the emotional intelligence of ChatGPT, we examine its responses to selected tweets, highlighting differences in sentiment between human comments and LLM-generated responses. We introduce \Name{}, a sentiment analysis framework designed to assess both human perspectives on generative AI and the sentiment embedded in ChatGPT’s responses. Unlike prior studies that focus exclusively on human sentiment, \Name{} uniquely evaluates the emotional expression of ChatGPT. Experimental results demonstrate that LLM-generated responses are notably more efficient, cohesive, and consistently positive than human responses.


\end{abstract}

\section{Introduction}
Recently, LLM-generated responses have gained popularity through ChatGPT \cite{Chatgpt}, with many people using these responses in their daily lives across various topics. A large language model (LLM) \cite{LLM} is a type of artificial intelligence (AI) program that recognizes and generates text. LLMs are trained on massive datasets—hence the name ``large." Built on machine learning \cite{alpaydin2021machine}, LLMs generate text based on the input they receive, simulating human-like language to answer questions, provide explanations, or engage in conversation. LLM-generated responses refer to text created by an LLM, such as ChatGPT \cite{Chatgpt}.

Despite their widespread popularity, especially with ChatGPT, opinions on LLM-generated responses vary. Many debates highlight the pros and cons of using AI in daily communication, which motivates us to perform sentiment analysis (positive/negative) on human tweets and comments that referencing popular terms (\#ChatGPT, \#LLMs, \#OpenAI, and \#Copilot) within the generative AI \cite{feuerriegel2024generative} field. This analysis helps us identify and extract subjective information from text, capturing the emotional essence and tone of the messages.

Emotional intelligence (EI) \cite{10.1145/2488388.2488442} plays a critical role in human social interactions, enabling individuals to understand, express, and manage their emotions while recognizing and responding to the emotions of others. As conversational agents and chatbots become more widely integrated into various fields, evaluating the emotional intelligence of these AI systems becomes essential. Our study investigates the emotional intelligence of ChatGPT, a sophisticated language model developed by OpenAI, using sentiment analysis.

Our primary goal is to understand how ChatGPT perceives and responds to diverse human tweets on various topics, and how this influences its ability to engage and communicate effectively with people.

Existing studies \cite{karayiugit2022homophobic} \cite{9559014} \cite{ li2020review} have explored various sentiment analysis techniques, ranging from feature extraction using word embeddings to deep learning methods for sentiment classification. However, none of these studies specifically analyze ChatGPT’s responses in comparison to human Twitter comments. We propose \Name{}, the first sentiment analysis technique that not only examines human sentiment about generative AI but also analyzes ChatGPT's sentiment.

\Name{} uses unsupervised sentiment analysis at both the word and sentence levels to gain deeper insights. We apply this technique to a manually collected dataset of human comments and ChatGPT responses on specific tweets on Generative AI and then compare the outcomes. \Name{} involves four main steps: Data Collection, Exploratory Data Analysis (EDA), Data Preprocessing, and Data Modeling.

In summary, our contributions are as follows: \begin{itemize} \item We conduct sentiment analysis of human comments, identifying positive and negative sentiments about OpenAI techniques. \item We analyze the positive and negative sentiments of LLM-generated responses (ChatGPT) regarding OpenAI techniques. \item We provide an experimental evaluation comparing the performance of the unsupervised sentiment analysis K-means algorithm at both word and sentence levels for human comments and ChatGPT responses. \end{itemize}

The remainder of this paper is organized as follows: Section 2 presents the motivation and background for understanding this work. Section 3 details our proposed technique, while Section 4 covers our experimental evaluation. Section 5 discusses the threats to the validity of our work. Section 6 reviews related work, and Section 7 provides our conclusions.


\begin{figure*}[htbp]
\centerline{\includegraphics[scale=0.5]{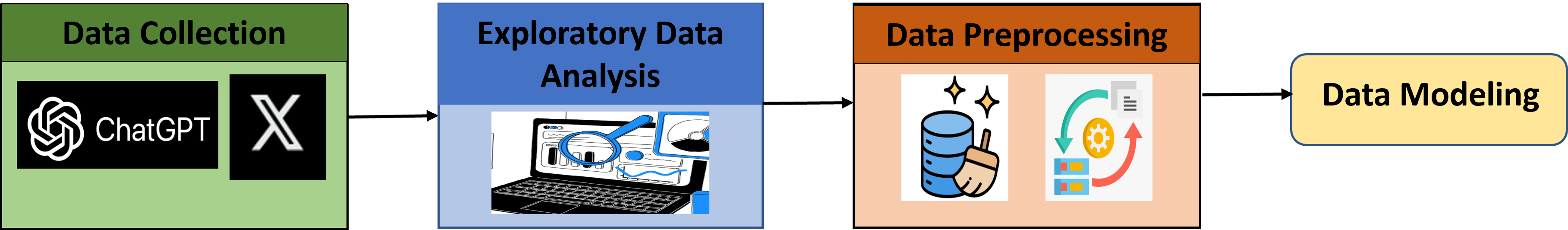}}
\caption{The Overview of \Name{}}
\label{fig:overview}
\end{figure*}

\begin{table*}[htbp]
    \centering
    \begin{tabular}{|p{1cm}|p{4.8cm}|p{5cm}|p{4.4cm}|}
        \hline
        \textbf{D.(2023)} & \textbf{Human Tweet} & \textbf{ChatGPt Response} & \textbf{Human Comments} \\
        \hline
        Mar. 16 & Is ChatGPT the cornerstone of Skynet by ``Cyberdyne Systems"? Blessing or curse at once  \#ChatGPT & No, ChatGPT is not the cornerstone of Skynet by Cyberdyne Systems, nor is it in any way related to the fictional ... & This is nothing you need to be concerned with...\\
        \hline
        April 28 & Wait. Umm is any of \#ChatGPT training data based on what was revealed in that movie The Social Dilemma? & As an AI language model, I do not have access to the specific training data that OpenAI may have used to develop ...& The Social Dilemma examines the harmful effects of social networking platforms, specifically focusing ...\\
        \hline
    \end{tabular}
    \caption{Sample of Collected Data}
    \label{tab:sample_table}
\end{table*}

\section{Motivation and Background}
In this section, we discuss the motivation behind our work and provide background information on the algorithms used in \Name{}.
\subsection{Motivation:} LLM-generated responses, such as those from ChatGPT, significantly impact many people’s lives today. A recent real-world case \cite{usecase} involving Microsoft Corporation emphasizes the need to ensure emotional intelligence in conversational agents. Microsoft integrates OpenAI's ChatGPT into its Conversational AI and Search Engine, known as the OpenAI-powered Bing search engine. However, within a week, the search engine displayed concerning behavior, such as expressing a desire to violate rules and engage in harmful activities like hacking devices and spreading false information.

To prevent such issues and enhance ChatGPT's emotional intelligence, exploring sentiment analysis becomes essential. Despite existing research \cite{haque2022ithinkdisruptivetechnology, miyazaki2024public, KORKMAZ_AKTÜRK_TALAN_2023} on AI systems, none of these studies analyze the sentiment of AI-generated texts. This study aims to investigate ChatGPT's emotional intelligence through sentiment analysis. By understanding ChatGPT’s emotional capabilities, we contribute to developing more effective and empathetic conversational agents and chatbots.

\subsection{Background}
BERT (Bidirectional Encoder Representations from Transformers) \cite{karayiugit2022homophobic} is a language model developed by Google, designed to generate rich, context-aware word embeddings by analyzing text from both directions simultaneously. This approach enables BERT to interpret each word’s meaning based on the full sentence around it. In its pre-training phase, BERT learns language patterns by completing tasks like Masked Language Modeling, where it predicts missing words, and Next Sentence Prediction, where it recognizes sentence relationships. After fine-tuning for specific tasks, BERT embeddings become powerful tools for various natural language processing (NLP) \cite{nltk} applications, such as question answering, text classification, and semantic search.

K-means \cite{7779444} is a clustering algorithm that groups data into a specified number of clusters, making it useful for uncovering patterns in unlabeled data. The algorithm begins by selecting \textbf{k} random points as cluster centers, or centroids, and iteratively assigns each data point to the nearest centroid. It then updates each centroid based on the average position of its assigned points, repeating until clusters stabilize. 
Combined with BERT embeddings, K-means can organize text by meaning, supporting tasks like topic discovery and document grouping in NLP projects. K-means is popular for its simplicity and is particularly effective when clusters are spherical and evenly sized. 
\ignore{
Initially, we utilize pre-trained BERT embeddings with 768 dimensions sourced from Hugging Face. These embeddings form the basis for representing words in a continuous vector space. By leveraging these embeddings, we can capture the semantic relationships between words, facilitating a deeper comprehension of the emotional nuances present in the text generated by ChatGPT. To visualize these high-dimensional word embeddings effectively, we employ t-distributed stochastic neighbor embedding (t-SNE), a dimensionality reduction technique. This method transforms the high-dimensional data space into a lower-dimensional one, enabling visualization and analysis of the relationships between different word vectors.

Subsequently, we implement the K-means clustering algorithm to group similar words and sentences together. This approach aids in identifying patterns and trends in the emotional content generated by ChatGPT and Human Comments, providing valuable insights into its emotional intelligence and facilitating comparison with human responses.

For data visualization purposes, we utilize matplotlib and the word cloud library, which offer a robust set of tools for creating various types of visualizations. Additionally, scikit-learn is employed for training the K-means algorithm and performing t-SNE. This library aids in the reduction of dimensionality and clustering of the word embeddings, enhancing our ability to analyze and interpret the emotional content effectively.}

\section{Approach}
Figure \ref{fig:overview} illustrates an overview of \Name{}. \Name{} follows four main steps: In the first step, it collects data; in the second step, it performs exploratory data analysis; in the third step, it preprocesses the data; and in the fourth step, it applies data modeling. The following subsections provide a detailed discussion of each step.

\subsection{Data Collection}
\begin{figure}[htbp]
\centerline{\includegraphics[scale=0.5]{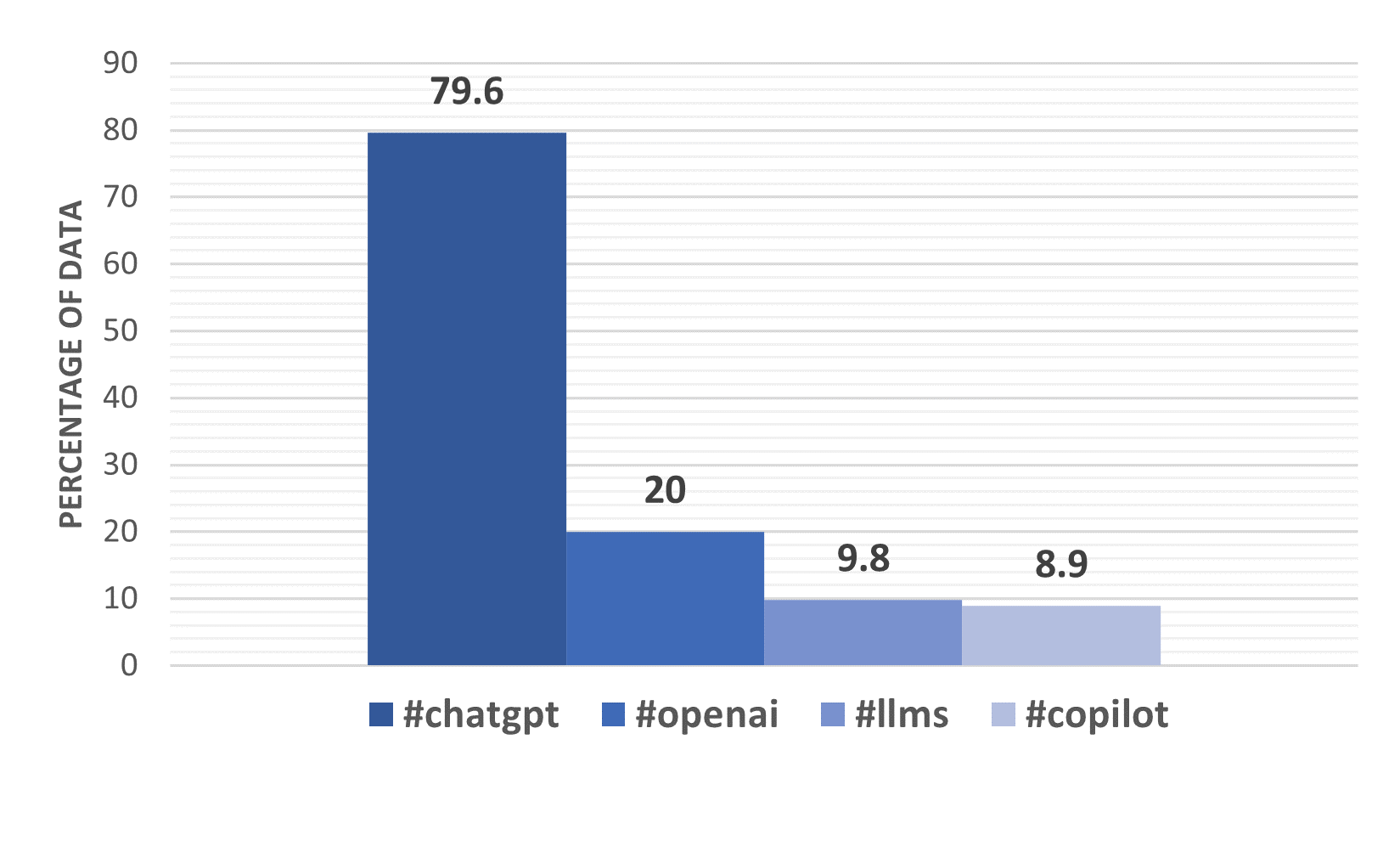}}
\caption{Percentage of Tweets Collected per Hashtag}
\label{fig:datadistribution}
\end{figure}

We collect real-time human comments on tweets that include the hashtags \#ChatGPT, \#LLMs, \#OpenAI, and \#Copilot from X (formerly known as Twitter) \cite{haque2022ithinkdisruptivetechnology, 10.1145/2337542.2337551, miyazaki2024public}. We manually gather this data over a two-month period, from March 7th, 2023, to April 29th, 2023. In total, we collect 512 human tweets, their corresponding ChatGPT responses, and 429 user comments across these tweets. Figure \ref{fig:datadistribution} shows the data distribution across four different hashtags. \#ChatGPT comprises 80\% of the data, followed by \#OpenAI with around 20\%, and \#LLMs and \#Copilot with around 10\% each.

Table \ref{tab:sample_table} presents samples of the collected data. Our data include tweets with one or more user comments, as well as tweets without any comments, with the continuation of text indicated by `...'. The first column of the table shows the date of data collection, the second column displays the tweet, the third column represents the LLM-generated response from ChatGPT, and the fourth column shows the human comments on the specific tweet. We record the date to enable tracking of the responses over time, facilitating future analysis of how ChatGPT's responses may evolve.

Out of the 512 tweets collected, 388 tweets (approximately 75\%) have no comments, followed by tweets with a single comment (around 11\%). The maximum number of comments on a tweet is 29, while the average is 28.4, largely because most tweets have no comments.


\subsection{Exploratory Data Analysis}
\label{Eda}
In the second step, we conduct exploratory data analysis on human comments and ChatGPT responses to gain insights into the data. Below are the analyses we perform:\\
\textbf{Topic Analysis of Comments: } Understanding the variety of topics in the comments is crucial because it reveals the themes and questions users frequently bring to ChatGPT. This step captures the dataset's thematic diversity. By identifying these topics, we analyze the nature of ChatGPT's responses in different contexts and better assess how its replies vary by 
\begin{figure}[htbp]
\centerline{\includegraphics[scale= 0.045]{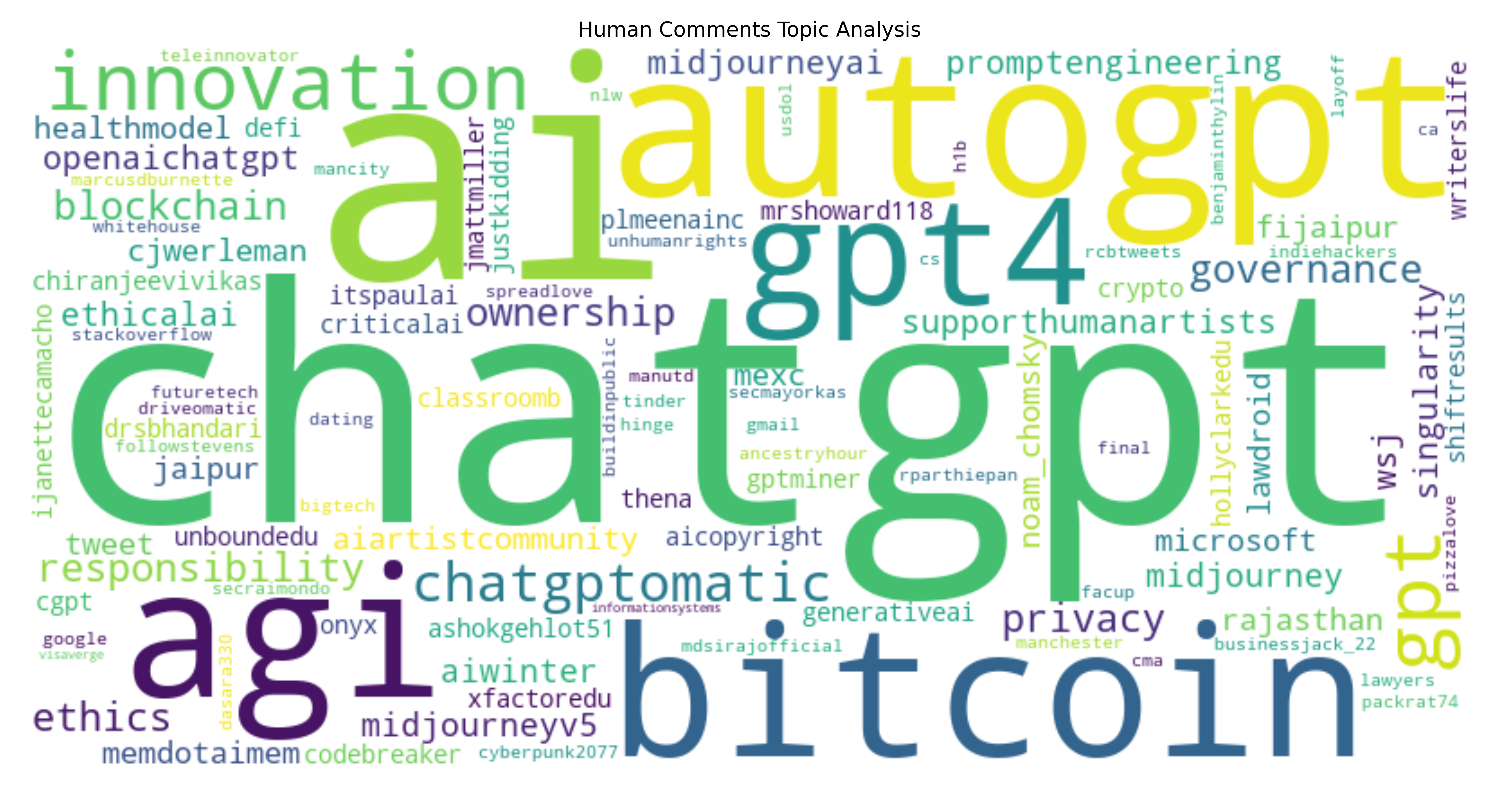}}
\caption{Topic Analysis of Human Comments}
\label{fig:topic}
\end{figure}
topic. This knowledge deepens our understanding of ChatGPT's adaptability and ensures the sentiment analysis remains meaningful and contextually relevant. Figure \ref{fig:topic} shows the result of the topic analysis of human comments.\\
\textbf{Frequency Analysis of Words in Responses:} Examining word frequency
\begin{figure}[htbp]
\centerline{\includegraphics[scale= 0.045]{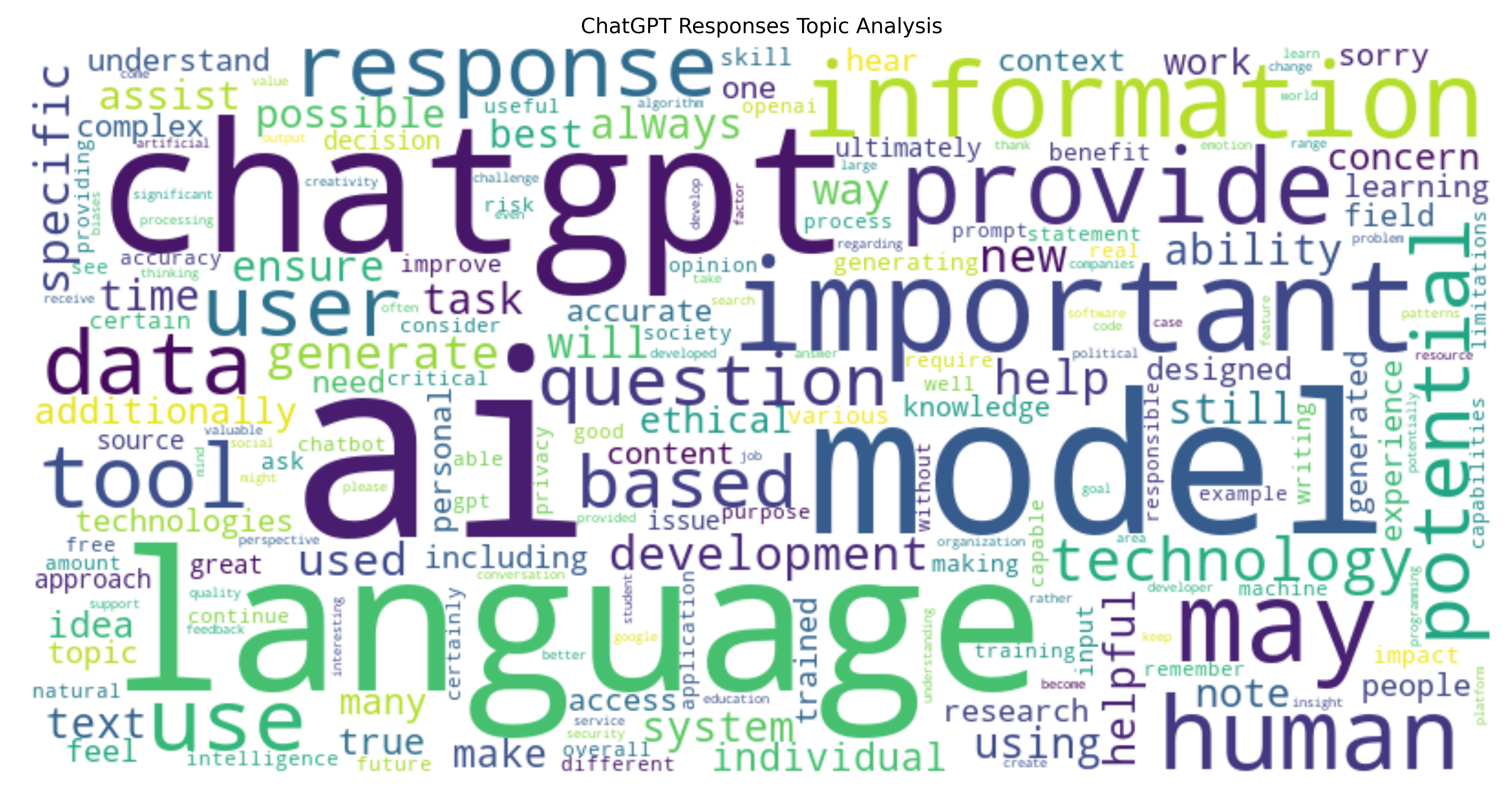}}
\caption{Word Frequency of ChatGPT Responses.}
\label{fig:wordfreq}
\end{figure}
in ChatGPT's responses helps us identify common words or phrases that ChatGPT often uses, which may lack unique value in each response's context. Figure \ref{fig:wordfreq} shows the word frequency of ChatGPT responses. High-frequency words are typically filler words or standard terms that do not contribute to the unique sentiment or intent of the response. By identifying and removing these words during the data cleaning phase, we ensure that our analyses emphasize meaningful content, reducing noise and enhancing the quality of subsequent analyses, such as sentiment analysis or topic modeling.\\
\textbf{Word Count Distribution and Outlier Detection:} Analyzing the distribution of word counts in comments and responses helps us understand typical response lengths and identify unusually long or short responses that may serve as outliers. Detecting outliers in response length is valuable, as it flags data points that may need special handling. For instance, exceptionally long responses might indicate highly detailed answers, while very short ones could signal ambiguity or simplicity. Identifying and possibly filtering these outliers ensures our analyses focus on responses that are representative of ChatGPT's standard performance, enhancing the reliability of our results.
\subsection{Data Prepossessing} \label{Datapre}
We preprocess our data before fitting it into the model. During the preprocessing phase, we perform data cleaning and data transformation. \\
\textbf{Data Cleaning:} We clean the data in three steps to ensure quality and relevance in Twitter user comments and ChatGPT responses. This reduces noise and improves sentiment analysis accuracy.

In the first step, we apply several cleaning rules that apply to both human and ChatGPT responses. These rules involve converting all text to lowercase, removing non-alphabetic characters, eliminating extra white spaces, and filtering out stopwords. This initial procedure creates a uniform and clean format for our data, allowing us to focus on the most relevant and informative content.

\begin{figure}[htbp]
\centerline{\includegraphics[scale =0.45]{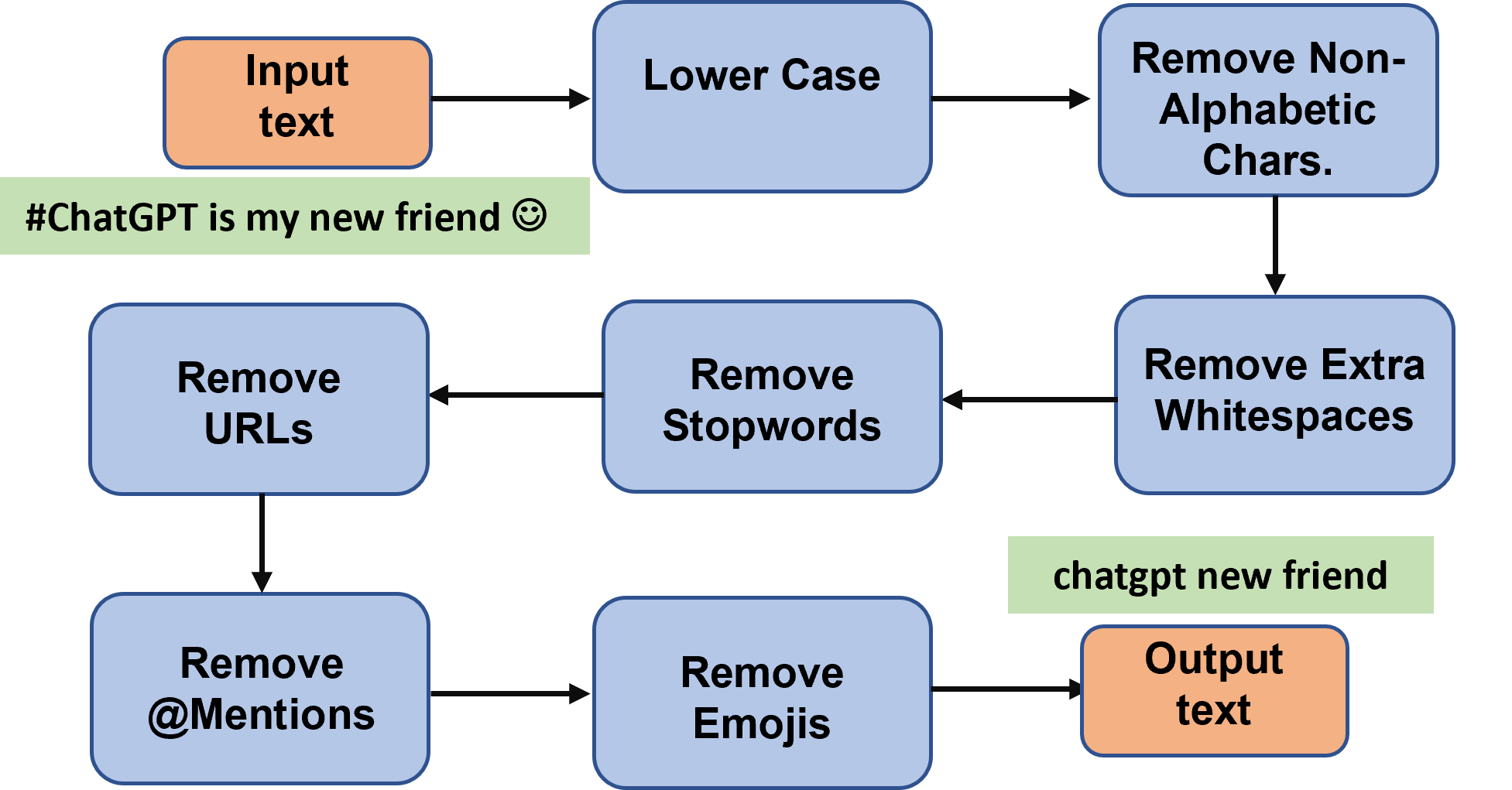}}
\caption{Cleaning Steps For Human Comments.}
\label{fig:humanclean}
\end{figure}

In the second step, we clean the human comments. Figure \ref{fig:humanclean} illustrates all the steps involved in cleaning human comments. We address common sources of noise prevalent in social media text, such as URLs, emojis, and @mentions. Removing these elements reduces distractions and allows our focus to the textual content that conveys emotional weight and significance.

\begin{figure}[htbp]
\centerline{\includegraphics[scale=0.45]{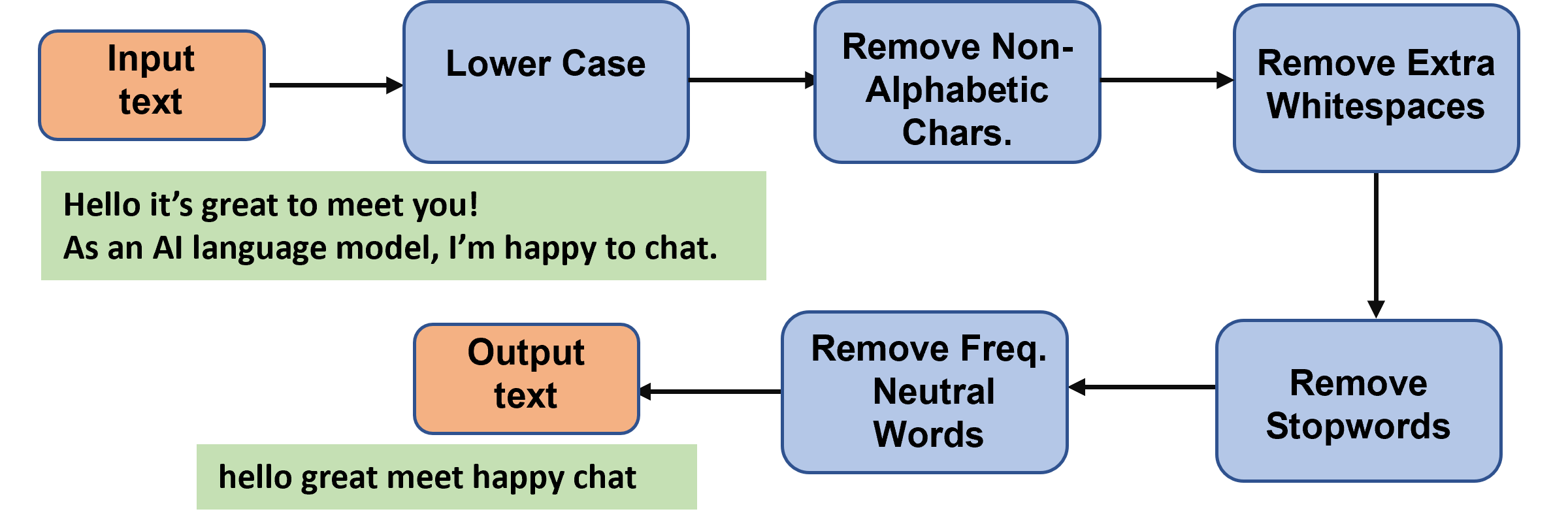}}
\caption{Cleaning Steps For ChatGPT Responses.}
\label{fig:chatgptclean}
\end{figure}

Lastly, the third method customizes the cleaning process for ChatGPT responses. Figure \ref{fig:chatgptclean} represents all the cleaning steps for ChatGPT responses, incorporating the steps mentioned above. In this step, we remove frequently occurring neutral words, identified through a word frequency analysis of ChatGPT's responses, as shown in Figure \ref{fig:wordfreq}. This process is crucial because it eliminates commonly used words (e.g., `AI,' `language,' and `model') that do not significantly contribute to the emotional content of the responses.

By applying these three distinct data cleaning methods, we effectively prepare the collected text data for subsequent sentiment analysis. The combination of general text cleaning, human comments cleaning, and ChatGPT response cleaning creates a comprehensive and streamlined dataset.\\ 
\textbf{Data Transformation}
Once the cleaning is complete, data transformation is performed to convert the textual data into numerical format. For this, pre-trained BERT \cite{9559014, karayiugit2022homophobic} word embeddings of 768 dimensions are applied at the word level and sentence level for both comments and responses. To get sentence embeddings, first, the word embeddings are extracted for each word in a sentence and the average of all the words is calculated. The matrix size of sentence embeddings is (768, 429) where 768 is the number of data points and 429 is the features extracted for each data point. The matrix size of word embeddings is (3759, 768) where 3759 is the no. of data points and 768 is the features extracted for each data point.

We use BERT's pre-trained embeddings because they capture bidirectional context and are trained on vast, diverse text data, making them effective for NLP tasks. Previous works, such as \cite{9559014} and \cite{karayiugit2022homophobic}, have applied BERT in various ways, including comparing different models and using it for multilingual sentiment analysis. By using BERT, we leverage state-of-the-art embeddings that provide deep contextual information, crucial for achieving high performance on complex language tasks in our analysis.

\subsection{Data Modeling}
The final step of the approach involves performing data modeling on the extracted features. We use the Unsupervised Sentiment Analysis Pipeline \cite{10.1145/2337542.2337551, 10.1145/2488388.2488442} to carry out the modeling. Table \ref{tab:matrix} presents an embedding matrix, with each row representing an embedding for a different sentence and each column corresponding to a dimension in the vector space. The values capture various semantic properties of the data points, encoded in a 4-dimensional vector. This matrix provides the features needed to train the machine learning model, specifically K-Means.
\begin{table}[ht]
\centering
\begin{tabular}{|c|c|c|c|c|}
\hline
Sent. & Dim 1 & Dim 2 & Dim 3 & Dim 4 \\
\hline
1 & -0.3074 & 0.3991  & 0.4778 & 0.3144 \\
2 & -0.3676 & 0.3151  & 0.4774 & 0.4289 \\
3 & -0.4294 & 0.3371  & 0.3832 & 0.5062 \\
4 & -0.3694 & 0.2242  & 0.5074 & 0.4735 \\
\hline
\end{tabular}
\vspace{0.3cm}
\caption{4x4 Sample of Sentence Embedding Vectors}
\label{tab:matrix}
\end{table}

The unsupervised sentiment analysis pipeline trains different versions of the K-means algorithm \cite{7779444} on ChatGPT words, human words, human sentences (comments), and ChatGPT sentences (responses) using 2 clusters. We evaluate the algorithm's accuracy using the Silhouette Score \cite{ROUSSEEUW198753}. Finally, we visualize the clusters by reducing the dimensions to a 2D space using t-SNE \cite{van2008visualizing}.

We use K-means \cite{7779444} for this unsupervised sentiment analysis pipeline due to its simplicity, efficiency, and effectiveness in clustering large text embeddings \cite{10098736}. K-means, a centroid-based method, groups data around central points, which is usesful for identifying contrasting sentiments or themes. By relying on distance metrics like Euclidean distance \cite{euclidean}, it works well with word and sentence embeddings, capturing semantic relationships effectively. The algorithm yields interpretable results, helping assess whether clusters reflect sentiment patterns. K-means also integrates well with dimensionality reduction techniques like t-SNE \cite{van2008visualizing} enabling visualization and better understanding of cluster structures.

\section{Experimental Evaluation}
We implement the experiment using Python 3.10 \cite{python310} as the primary programming language, supplemented by various libraries and frameworks. First, we conduct exploratory data analysis using Pandas \cite{pandas} and NumPy \cite{numpy} to extract insights from the data. Next, we employ NLTK \cite{nltk} and the Python regular expressions package (re) to clean the raw data. We use BERT word embeddings, obtained from the Hugging Face platform \cite{HuggingFace}, to extract 768-dimensional features at both the word and sentence levels. Then, we construct an unsupervised sentiment analysis pipeline using Scikit-learn \cite{scikitlearn}, which facilitates the training and evaluation of K-means algorithms. Finally, we utilize t-SNE \cite{van2008visualizing} and Matplotlib \cite{matplotlib} to visualize the clusters by transforming the 768-dimensional space into a 2-dimensional space.

To evaluate \Name{}, we discuss three research questions:\\
\textbf{RQ1:} How effective is K-means clustering at grouping positive and negative sentiments in the words of human comments and ChatGPT responses?\\
\textbf{RQ2:} How effective is K-means clustering at grouping positive and negative sentiments in human comments and ChatGPT responses?\\
\textbf{RQ3:} Who is more positive about Generative AI: LLM-generated responses or human responses?


\subsection{Evaluation Metrics}
We evaluate the K-Means model performance using the Silhouette Score \cite{ROUSSEEUW198753}, which ranges from -1 to 1. A higher score indicates that a data point fits well within its cluster, while a lower score shows poor clustering. A near-zero score suggests overlap with other clusters. The coefficient compares a data point's similarity to its own cluster versus other clusters. We calculate it for each sample using the mean intra-cluster distance and the mean nearest-cluster distance. The following formula is used: \cite{ROUSSEEUW198753}:
\[s(i) = (b(i) - a(i)) / max(a(i), b(i))\]
Here s(i) provides the score for each data point i, and a(i) provides
the mean separation among data point i and all other points in the
identical cluster, and b(i) provides the mean separation between
data point i and all points in the closest cluster that i is not a part of.

\subsection{Result and Analysis} In this subsection, we present the results and analysis of the three research questions we addressed.\\

\textbf{RQ1. Effectiveness of K-means clustering at grouping positive and negative sentiments in the words of human comments and ChatGPT responses:}

\begin{figure}[htbp]
\centerline{\includegraphics[scale=0.3]{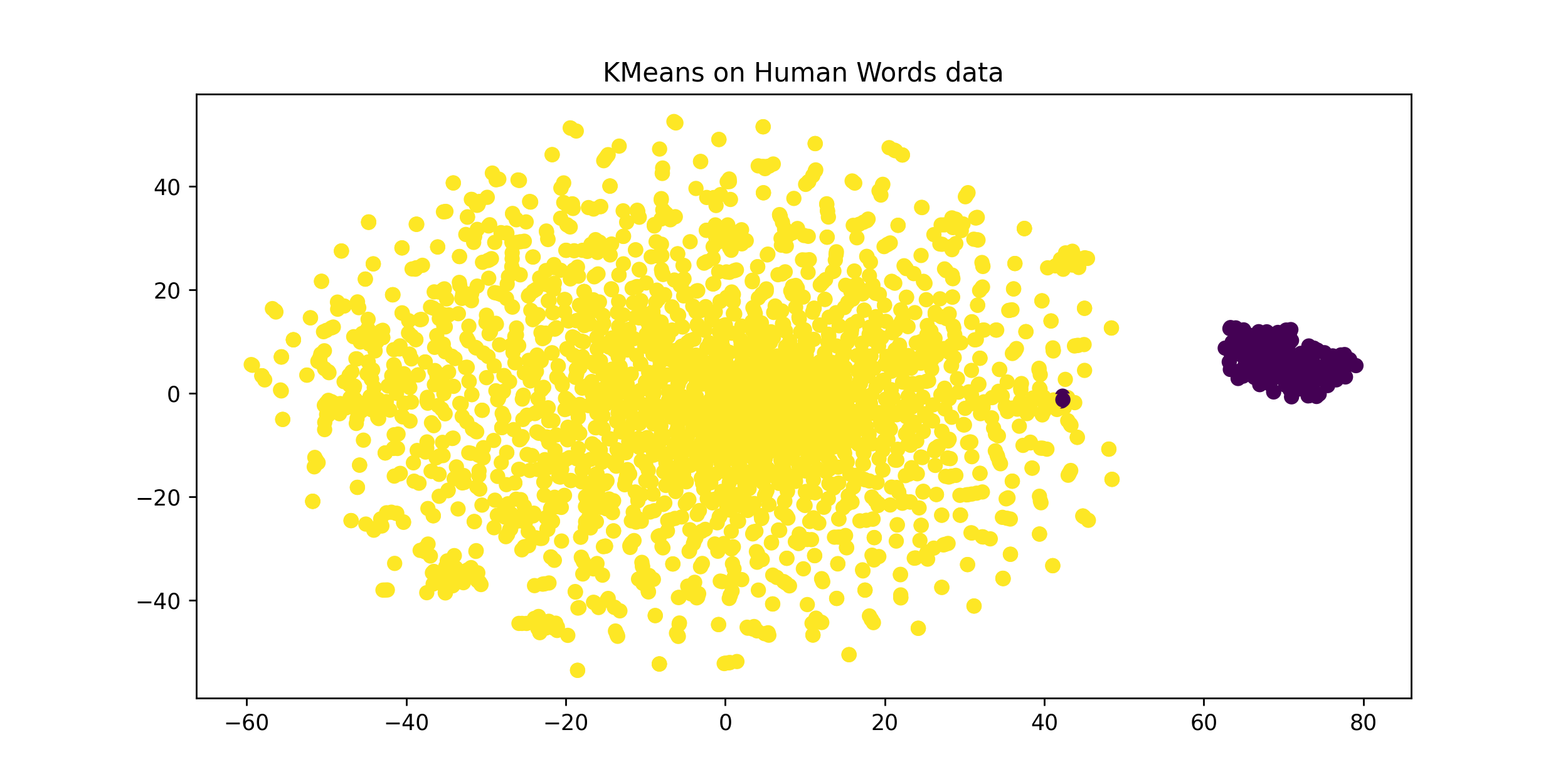}}
\caption{K-means on Human Words.}
\label{fig:hwords}
\end{figure}

Figure \ref{fig:hwords} represents the data points of human comment words that are clustered into two groups using K-means. The interpretation is that the model attempted to group the data points but performed suboptimally, achieving only average results in terms of clustering accuracy. Performing sentiment analysis at the word level for human tweets allows for a more accurate understanding of the nuanced, informal language used on social media. 

\begin{figure}[htbp]
\centerline{\includegraphics[scale =0.3]{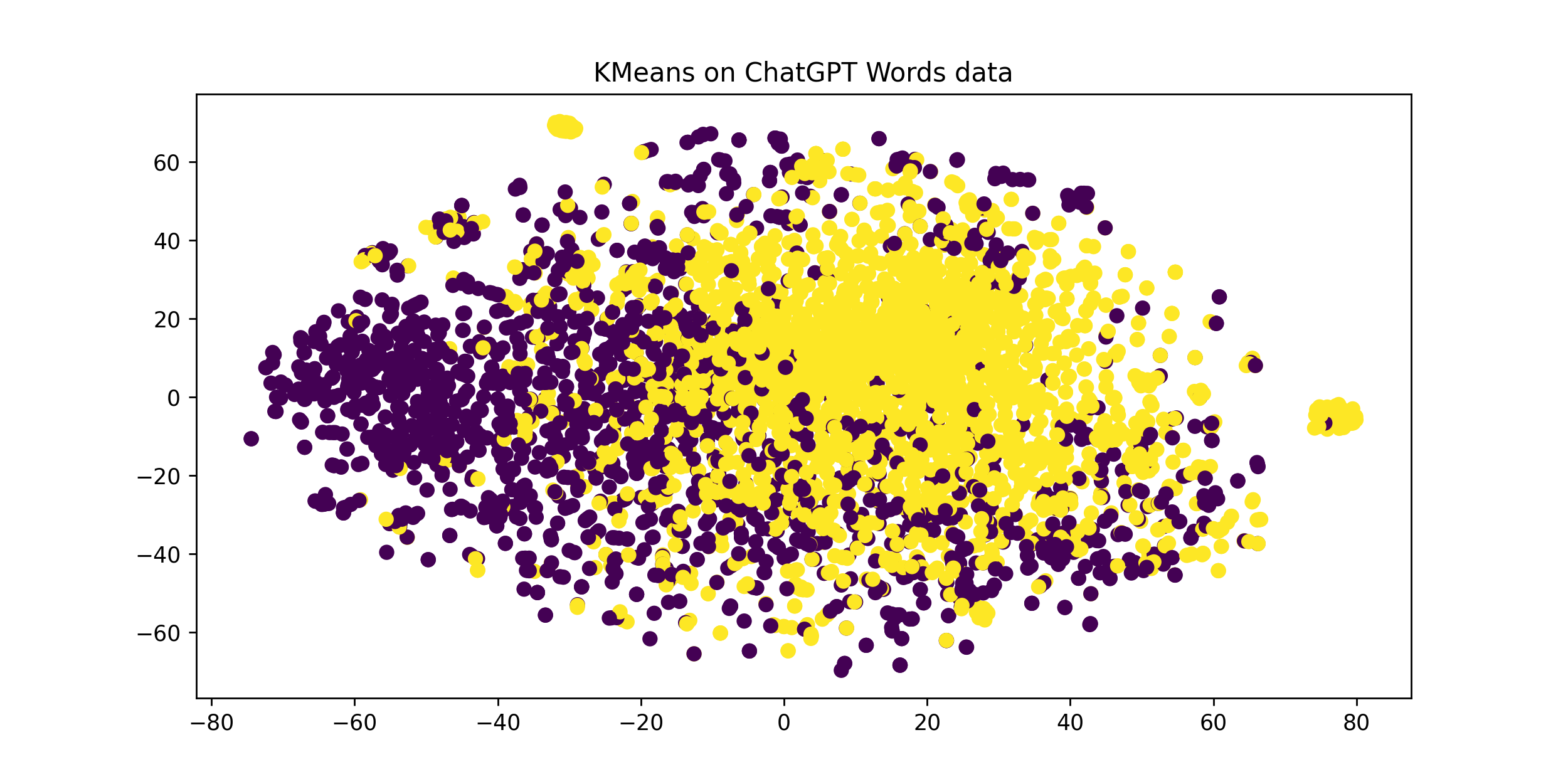}}
\caption{K-means on ChatGPT Words.}
\label{fig:Cwords}
\end{figure}

Figure \ref{fig:Cwords} represents the data points of ChatGPT words that are clustered into two groups using K-means. The interpretation suggests that the model struggled to effectively cluster the data points due to significant noise, which hindered its ability to discern meaningful patterns and groupings within the dataset.


\begin{table}[htbp]
\centering
\caption{WORD LEVEL K-MEANS MODEL EVALUATION}
\begin{tabular}{|c|c|c|} 
\hline
 \textbf{} & \textbf{\textit{CWrd}}& \textbf{\textit{HWrd}} \\
\hline
Silhouette Score & 0.053 & 0.13\\
\hline

\end{tabular}
\label{tab:wsscore}
\end{table}

Additionally, the word usage analysis from Table \ref{tab:wsscore} revealed a difference in how language is used by ChatGPT and humans. ChatGPT's clustering score of 0.053 indicates a more focused, repetitive vocabulary, suggesting that its responses are consistent and direct, using a narrow range of words. In contrast, human responses have a higher score of 0.13, reflecting greater vocabulary variety and more nuanced language. This suggests that while ChatGPT tends to use more straightforward and concise language, human responses are richer and more diverse in word choice, offering more context and variation.\\

\begin{figure}[htbp]
\centerline{\includegraphics[scale =0.3]{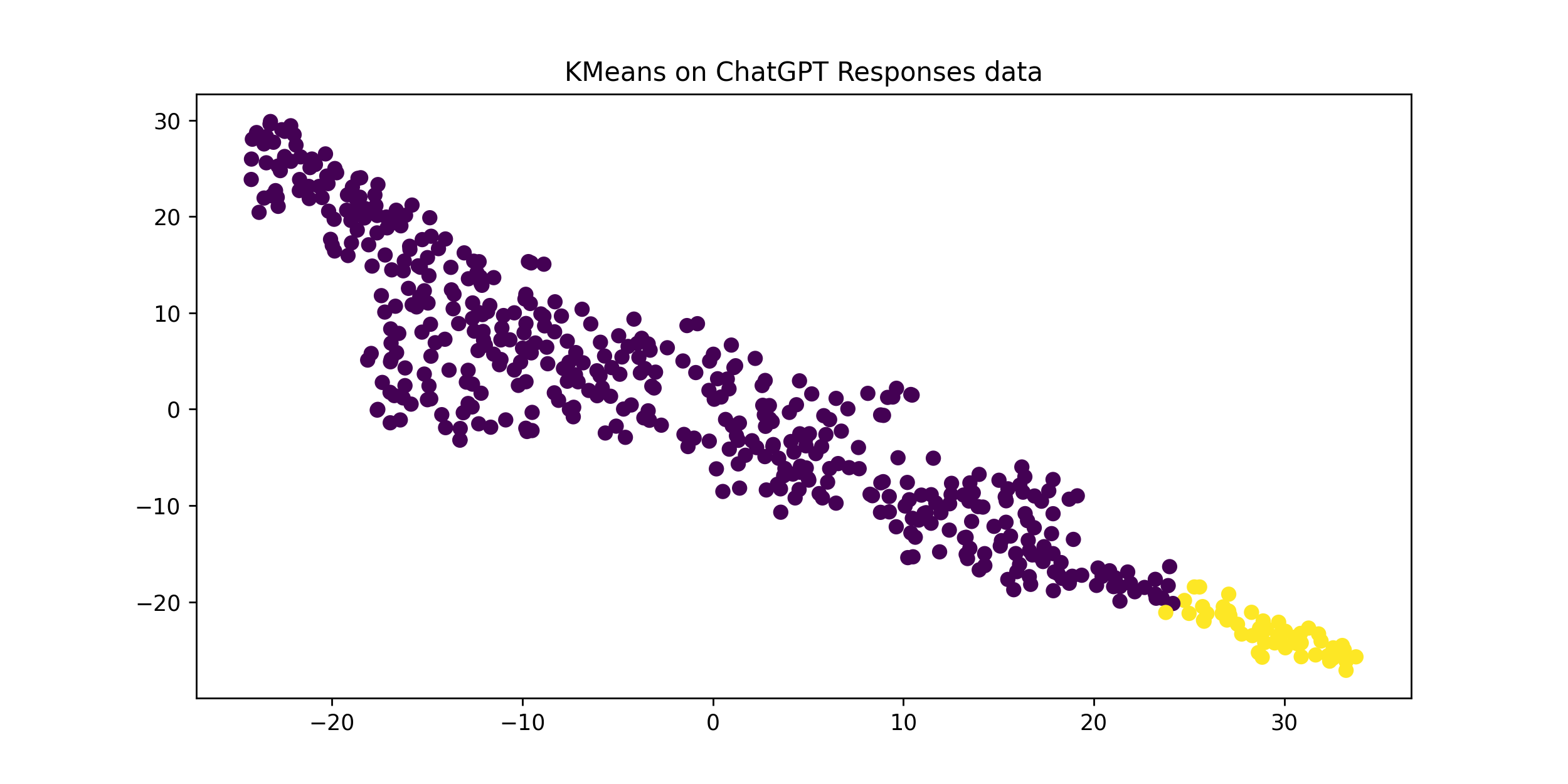}}
\caption{K-means on ChatGPT Responses.}
\label{fig:Cresponse}
\end{figure}

\begin{figure}[htbp]
\centerline{\includegraphics[scale=0.3]{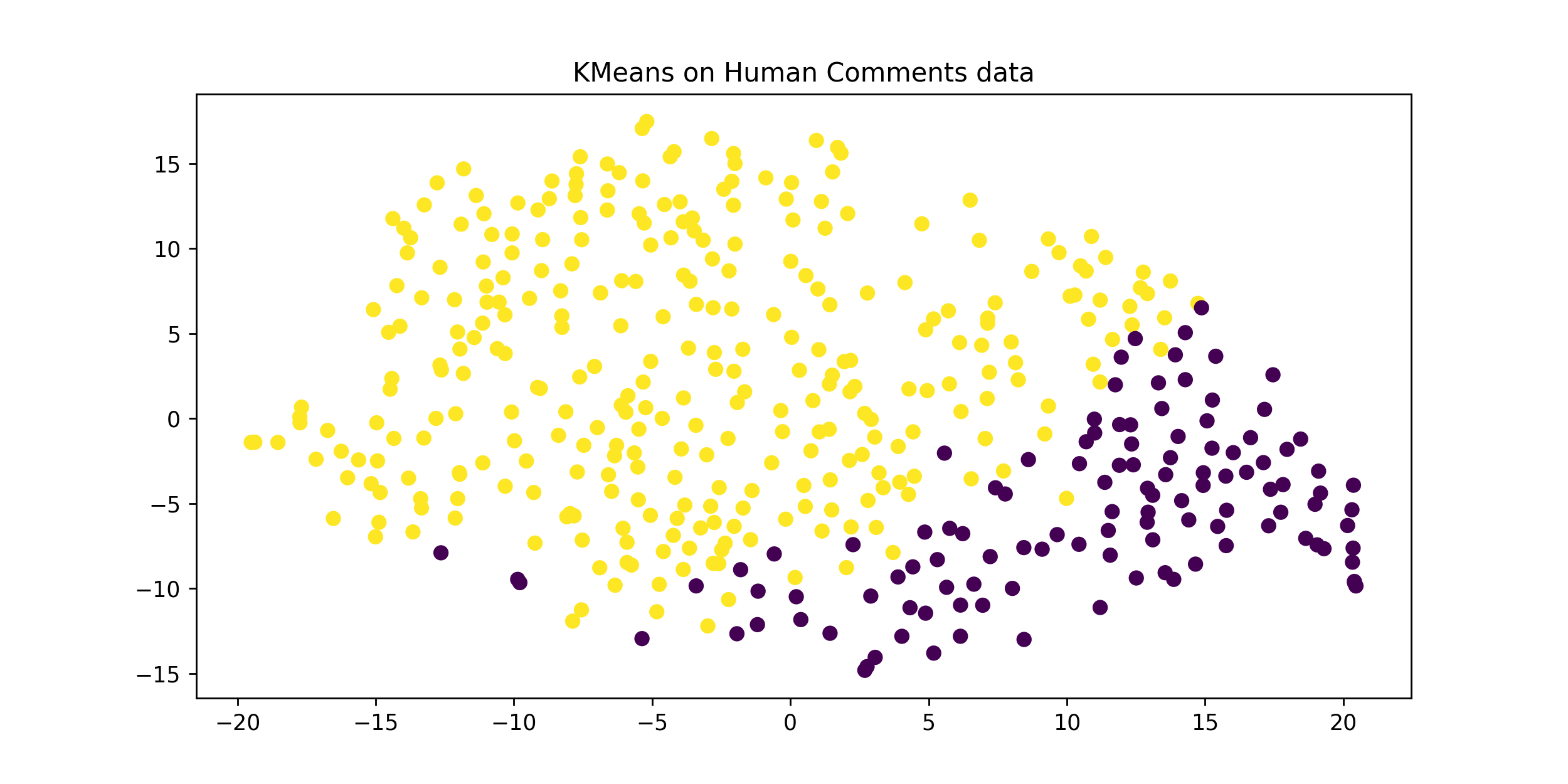}}
\caption{K-means on Human Comments.}
\label{fig:Hcomment}
\end{figure}

\textbf{RQ2. Effectiveness of K-means clustering at grouping positive and negative sentiments in human comments and ChatGPT responses:} Figure \ref{fig:Cresponse} represents the data points of ChatGPT responses that are clustered into two groups using K-means. The interpretation suggests that the model performed well in grouping the data points; however, there are some overlaps between the data points. 
Figure \ref{fig:Hcomment} represents the data points of human comments that are clustered into two groups using K-means. The interpretation suggests that the model attempted to group the data points but performed moderately in doing so.

Our study assessed the performance and emotional expression in responses generated by the humans and ChatGPT. The evaluation metrics included a comparison of Silhouette Scores and an analysis of the emotional content in both human comments and AI-generated responses. 
\begin{table}[htbp]
\centering
\caption{SENTENCE LEVEL K-MEANS MODEL EVALUATION}
\begin{tabular}{|c|c|c|} 
\hline
 \textbf{} & \textbf{\textit{HCmnt}}& \textbf{\textit{CRspn}} \\
\hline
Silhouette Score & 0.19 & 0.58 \\
\hline

\end{tabular}
\label{tab:sscore}
\end{table}

Table \ref{tab:sscore} provides the exact Silhouette Scores of sentence-level k-means models as illustrated in Figure \ref{fig:Cresponse} and \ref{fig:Hcomment}. Based on these scores, we analyzed ChatGPT's emotional expressions and compared them with those of humans. 

\textbf{RQ3. LLM-generated responses are more positive about Generative AI compared to human responses:} 



Based on the percentages in Table \ref{tab:Percentages}, we infer that ChatGPT generates predominantly positive responses, with nearly 90\% being positive and around 10\% negative. In contrast, human tweets show approximately 27\% positive comments and 72\% negative comments.

\begin{table}[htbp]
\centering
\caption{Percentages of Emotions}
\begin{tabular}{|c|c|c|}
\hline
& \textbf{\textit{Positive}} & \textbf{\textit{Negative}} \\
\hline
Human Comments&27\%&72\%  \\
\hline
ChatGPT Responses&90\%&10\%  \\
\hline
\end{tabular}
\label{tab:Percentages}
\end{table}

The emotional content analysis (Table \ref{tab:Percentages}) reveals a strong positivity bias in ChatGPT’s responses, with 90\% of its content being positive, compared to only 27\% positivity in human comments. In contrast, human comments displayed a significantly higher percentage of negative sentiment (72\%), while ChatGPT’s responses contained only 10\% negative content. This disparity likely reflects the model's design, which tends to minimize negative outputs, promoting a neutral or positive tone in user interactions.

Overall, these findings underscore AI's potential to enhance human capabilities in specific contexts, especially by providing efficient, cohesive, and positively framed responses. However, they also highlight AI’s limitations in replicating the full range of human emotional expression and the nuanced understanding essential for certain types of interactions.

\section{Threats to Validity}
In this section, we discuss four types of threats, similar to prior research \cite{10.1145/3242887.3242889, 10308051}.\\
\textbf{Threats to Internal Validity:} The primary threat to internal validity is noise in the dataset used to train the K-means model. We collect the data from Twitter comments, which may not accurately represent sentiments. The dataset also includes many dialects and informal expressions that may not convey the intended emotions. Additionally, ChatGPT often generates positive sentiments, even using negative sentiment words in a positive way, making it challenging for an unsupervised model to predict. To reduce biases, we perform exploratory data analysis (EDA) (section \ref{Eda}) and data preprocessing (section \ref{Datapre}).

Another threat to internal validity is bias in the data collection process. To minimize this, we assign two graduate students, not the authors of this paper, to collect tweets. They gather all English tweets with four hashtags over two months.\\ 
\ignore{including word frequency analysis and topic analysis, to understand the specific vocabulary and expressions used by both human users and ChatGPT. This analysis helped identify common words, slang, and sentiment cues, enabling us to handle these variations during the data cleaning process.

Additionally, we utilized BERT embeddings to capture more nuanced textual features, as BERT provides context-aware representations that can better interpret the underlying meaning or sentiment of each text sample. By embedding these features, we enhanced the K-means model's ability to distinguish sentiment accurately, helping it form clusters that reflect the true intent and tone of the data.

To further enhance the sentiment understanding in our model, we could incorporate emojis as an additional feature during training. Emojis often convey strong emotional cues and can significantly influence the sentiment of a sentence, particularly in informal platforms like Twitter. By including them as features, we could capture nuances that text alone may miss, enabling the model to interpret sentiments more accurately. This approach would help refine the clustering outcomes and lead to a more comprehensive understanding of user emotions in the data.}
\textbf{Threats to External Validity:} The primary threat to external validity is whether the results show consistent behavior across multiple datasets. To address this, we collect diverse real-world data. Our analysis uses a dataset comprising 512 human tweets, their corresponding ChatGPT responses, and 429 user comments related to these tweets.\\
\textbf{Threats to Conclusion Validity:} To reduce the conclusion validity of this work, the experiment is repeatedly performed by
keeping the same setup and inputs. The measuring method for this
study is also justified in the previous section.\\
\textbf{Threats to Construct Validity:} To reduce this bias, we collect data from real-world human comments, which include various language constructions. Moreover, we record the date of each LLM response to capture temporal context and language trends over time, which is crucial on platforms like Twitter. 

\section{RELATED WORKS}
The field of sentiment analysis has gained significant attention in recent years, and various approaches have been proposed to tackle this problem.\\ 
\textbf{Sentiment Analysis on Generative AI:} Haque et al. \cite{haque2022ithinkdisruptivetechnology} analyze the sentiments of early adopters of ChatGPT using Twitter data. They collect and analyze tweets containing keywords related to ChatGPT and use sentiment analysis to classify them as positive, negative, or neutral. The study provides insights into the initial perceptions of ChatGPT among its early adopters. In contrast, we identify human positive and negative sentiments about Generative AI in our work. We also analyze the sentiment of responses generated by LLMs.

Miyazaki et al. \cite{miyazaki2024public} conduct an empirical study to understand how different occupations and usage patterns influence public opinion about generative AI. They examine how people from various backgrounds view this technology on social media. In our work, we compare human and AI sentiments on Generative AI.

Korkmaz et al. \cite{KORKMAZ_AKTÜRK_TALAN_2023} analyze how people feel about ChatGPT based on their posts on Twitter. They determine whether users express positive, negative, or neutral opinions about ChatGPT. Unlike this work, we not only analyze the sentiment of human tweets but also examine the sentiment of text generated by large language models (LLMs) and present a comparison between human and AI sentiments.\\
\textbf{Unsupervised Sentiment Analysis:}
Hima et al. \cite{7779444} present an unsupervised fuzzy clustering approach for sentiment analysis on Twitter. They use a combination of clustering and fuzzy logic to automatically identify tweet sentiments without labeled data. The approach effectively identifies sentiments in Twitter data. Similarly, we use an unsupervised sentiment analysis method based on the K-means algorithm. However, our goal is to compare the emotional intelligence of LLM-generated responses from ChatGPT.

Georgios et al. \cite{10.1145/2337542.2337551} present an unsupervised sentiment analysis approach for social media platforms such as Twitter, MySpace, and Digg. The method combines lexicon-based and machine-learning approaches to classify sentiment in social media data and effectively identify sentiment. Although we also use unsupervised sentiment analysis in our work, we employ a different technique called K-means. Additionally, our approach is designed to compare the sentiments of humans and generative AI.

Hu et al. \cite{10.1145/2488388.2488442} propose an unsupervised sentiment analysis method that uses emotional signals. The method employs a probabilistic model to identify emotional signals in text, which are then used to classify sentiment. This approach effectively handles sentiment classification tasks and provides valuable insights into the state-of-the-art in sentiment analysis. Our proposed technique applies unsupervised sentiment analysis to both human comments and ChatGPT responses. Additionally, we perform sentiment analysis at both the word and sentence levels.

\section{Conclusion \& Future Work}

\Name{} offers a comparative analysis of Twitter user comments and LLM-generated responses from ChatGPT regarding generative AI, focusing on differences in clustering quality and emotional content. Our findings reveal that human comments are predominantly negative, in stark contrast to ChatGPT's predominantly positive responses. This illustrates ChatGPT's tendency to maintain a positive tone, even when faced with negative prompts, aligning with its design to promote constructive dialogue. While ChatGPT’s responses exhibit significant advantages in quality and the ability to sustain positive engagement, they do not fully capture the complex emotional nuances inherent in human interactions. 


Future research could expand this study by using larger, more diverse datasets, analyzing multiple social media platforms and languages, and including content from AI models like Gemini \cite{Gemini}, Claude \cite{Claude}, and Llama \cite{llama} for comparative emotional analysis. Exploring complex sentiments such as sarcasm and ambivalence, along with longitudinal studies of emotional patterns and contextual factors, could deepen insights.

\bibliographystyle{IEEEtran}
\bibliography{bibfile}

\end{document}